\documentclass[11pt,a4paper]{article}
\usepackage[hyperref]{acl2020}
\usepackage{times}
\usepackage{latexsym}

\usepackage{microtype}

\usepackage{url}
\usepackage{enumitem}
\usepackage{booktabs}
\usepackage{multirow}
\usepackage{graphicx}
\usepackage{array}
\usepackage{subfigure}
\usepackage{amsmath}
\usepackage{makecell}
\usepackage{amsmath}
\usepackage{amsfonts}
\usepackage{amsthm,amssymb,amsopn,bm}
\usepackage{color,soul}
\usepackage{verbatim}
\usepackage{xcolor,colortbl}
\definecolor{green}{rgb}{0.1,0.1,0.1}
\usepackage{tabulary}
\usepackage{footnote}
\makesavenoteenv{algorithm}
\usepackage{tabularx,booktabs}
\newcolumntype{Y}{>{\centering\arraybackslash}X}
\usepackage{paralist}

\setlist{leftmargin=5mm}

\DeclareMathOperator*{\argmax}{argmax}

\usepackage{tikz}

\newcolumntype{x}[1]{%
>{\raggedleft\hspace{0pt}}p{#1}}%

\aclfinalcopy 

\providecommand{\sewon}[1]{
}

\providecommand{\luke}[1]{
    {\protect\color{magenta}{\bf [Luke: #1]}}
}

\newcommand\tf[1]{\textbf{#1}}

\newcommand\mf[1]{\mathbf{#1}}
\newcolumntype{P}[1]{>{\centering\arraybackslash}p{#1}}

\title{
    Knowledge Guided Text Retrieval and Reading \\
    for Open Domain Question Answering
}

\author{
    Sewon Min$^1$, Danqi Chen$^{2}$, Luke Zettlemoyer$^{1,3}$, Hannaneh Hajishirzi$^{1,4}$ \\
    \hspace{-10pt}$^1$University of Washington, Seattle, WA \hspace{20pt}
    $^2$Princeton University, Princeton, NJ \\
    $^3$Facebook AI Research, Seattle, WA \hspace{30pt}
    $^4$Allen Institute for AI, Seattle, WA \\
    {\tt $\{$sewon,lsz,hannaneh$\}$@cs.washington.edu}
    \quad {\tt danqic@cs.princeton.edu}
}

\date{}

\begin{document}
\maketitle
\newcommand{\none}{\texttt{no\_relation}}
\newcommand{\parent}{\texttt{parent}}
\newcommand{\child}{\texttt{child}}
\newcommand{\hyper}{\textit{hyperlink}}

\newcommand{\triviaqa}{\textsc{TriviaQA}}
\newcommand{\nq}{\textsc{Natural Questions}}
\newcommand{\webq}{\textsc{WebQuestions}}
\newcommand{\webqsplit}{\textsc{WebQuestions}}
\newcommand{\squad}{\textsc{SQuAD-open}}
\newcommand{\wikihop}{\textsc{WikiHop}}
\newcommand{\hotpot}{\textsc{HotpotQA}}

\newcommand{\tfidf}{TF-IDF}
\newcommand{\bert}{\textsc{BERT}}
\newcommand{\distilbert}{\textsc{DistilBERT}}

\newcommand{\parallelqa}{\textsc{ParReader}}
\newcommand{\parallelqabatch}{\textsc{ParReader++}}
\newcommand{\gretriever}{\textsc{GraphRetriever}}
\newcommand{\greader}{\textsc{GraphReader}}
\newcommand{\reader}{reader}
\newcommand{\readercap}{Reader}
\newcommand{\oldretrieval}{Text-match + Wikidata}
\newcommand{\sota}{state-of-the-art}
\newcommand{\dev}{development}
\newcommand{\redtext}[1]{{\protect\color{red}{#1}}}
\newcommand{\bluetext}[1]{{\protect\color{blue}{#1}}}


\newcommand{\webqpardev}{33.2}
\newcommand{\webqpartest}{33.0}
\newcommand{\webqbatchdev}{33.7}
\newcommand{\webqbatchtest}{31.8}
\newcommand{\webqbinarydev}{34.0}
\newcommand{\webqbinarytest}{36.4}
\newcommand{\webqreldev}{34.0}
\newcommand{\webqreltest}{36.0}

\newcommand{\nqpardev}{30.2}
\newcommand{\nqpartest}{29.3}
\newcommand{\nqbatchdev}{33.1}
\newcommand{\nqbatchtest}{33.5}
\newcommand{\nqbinarydev}{34.2}
\newcommand{\nqbinarytest}{34.1}
\newcommand{\nqreldev}{34.7}
\newcommand{\nqreltest}{34.5}

\newcommand{\triviaqapardev}{54.8}
\newcommand{\triviaqapartest}{54.7}
\newcommand{\triviaqabatchdev}{55.5}
\newcommand{\triviaqabatchtest}{55.0}
\newcommand{\triviaqabinarydev}{55.2}
\newcommand{\triviaqabinarytest}{54.2}
\newcommand{\triviaqareldev}{55.8}
\newcommand{\triviaqareltest}{56.0}


\newcommand{\webqscbatch}{29.4}
\newcommand{\webqscbinary}{30.8}
\newcommand{\nqscbatch}{30.5}
\newcommand{\nqscbinary}{32.6}


\newcommand{\webqfc}{33.7}
\newcommand{\webqempty}{33.7}
\newcommand{\webqcross}{34.2}
\newcommand{\webqinner}{33.4}
\newcommand{\nqfc}{33.6}
\newcommand{\nqempty}{33.6}
\newcommand{\nqcross}{33.7}
\newcommand{\nqinner}{33.7}


\newcommand{\webqelm}{32.9}
\newcommand{\webqbilinear}{32.6}
\newcommand{\nqelm}{33.8}
\newcommand{\nqbilinear}{34.0}
\newcommand{\triviaqaelm}{55.9}
\newcommand{\triviaqabilinear}{55.7}


\newcommand{\webqpair}{31.3}
\newcommand{\webqpairall}{25.7}
\newcommand{\nqpair}{29.8}
\newcommand{\nqpairall}{20.5}

\begin{abstract}
    We introduce an approach for open-domain question answering (QA) that retrieves and reads a passage graph, where vertices are passages of text and edges represent relationships that are derived from an external knowledge base or co-occurrence in the same article. Our goals are to boost coverage by using knowledge-guided retrieval to find more relevant passages than text-matching methods, and to improve accuracy by allowing for better knowledge-guided fusion of information across related passages. Our graph retrieval method expands a set of seed keyword-retrieved passages by traversing the graph structure of the knowledge base.
    Our reader extends a \bert-based architecture and updates passage representations by propagating information from related passages and their relations, instead of reading each passage in isolation.
    Experiments on three open-domain QA datasets, \webq, \nq\ and \triviaqa, show improved performance over non-graph baselines by 2-11\% absolute. Our approach also matches or exceeds the \sota\ in every case, without using an expensive end-to-end training regime.
\end{abstract}

\section{Introduction}\label{sec:intro}

\begin{figure}
\centering
\resizebox{\columnwidth}{!}{\includegraphics[width=\textwidth]{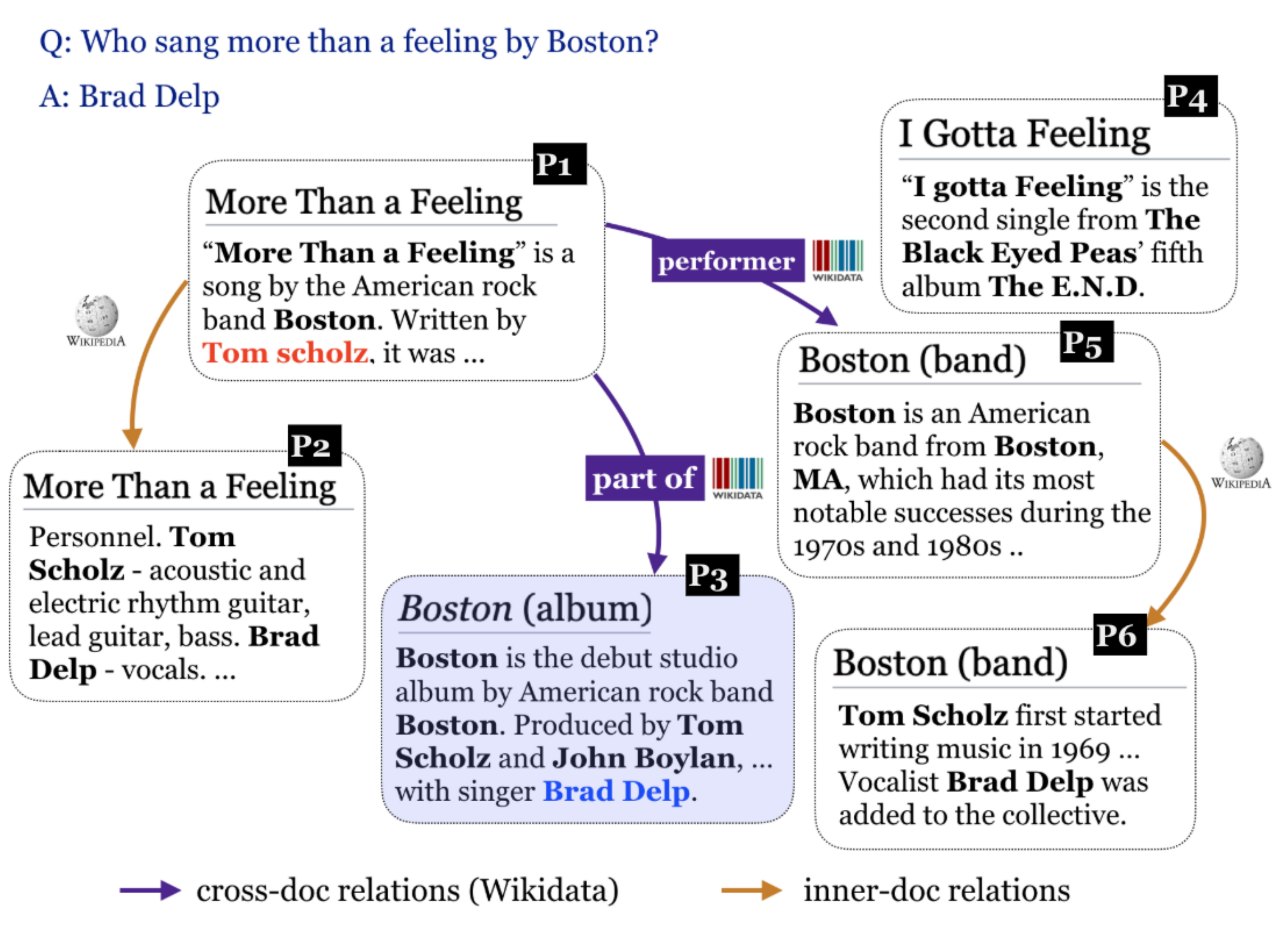}}
\caption{An example from \nq.
A graph of passages is constructed based on Wikipedia and Wikidata, where the edges represent relationships between passages.
The baseline model which uses \tfidf\ and reads each passage in isolation outputs the wrong answer (\redtext{red}) by selecting a person name from the passage about the song.
Our model which leverages relations (e.g. {\em part of}) synthesizes the context over related passages and predicts the correct answer (\bluetext{blue}) by choosing a singer of the album.
}
\label{fig:intro}\end{figure}

Open-domain question answering systems aim to answer any question a user can pose, with evidence provided by either factual text such as Wikipedia~\citep{chen2017reading,yang2019end} or knowledge bases (KBs) such as Freebase~\citep{berant2013semantic,kwiatkowski2013scaling,yih2015semantic}.
Textual evidence, in general, has better coverage but KBs more directly support making complex inferences. It remains an open question how to best make use of KBs without sacrificing coverage in text-based open domain QA. Previous work has converted KB facts to sentences to provide extra evidence~\citep{weissenborn2017dynamic,mihaylov2018knowledgeable}, but do not explicitly use the KB graph structure. In this paper, we show that such structure can be highly beneficial for both retrieving text passages and fusing information across them in open-domain text-based QA, for example as shown in Figure~\ref{fig:intro}.

We introduce a general approach for text-based open-domain QA that is knowledge guided: it retrieves and reads a passage graph, where vertices are passages of text and edges represent  relationships that are derived either from an external knowledge base or co-occurrence in the same article.
Our goal is to combine the high coverage of textual corpora with the structural information in knowledge bases, to improve both the retrieval coverage and accuracy of the resulting model.
Unlike standard approaches that retrieve and read a set of passages~\cite{chen2017reading}, our approach integrates graph structure at every stage to construct, retrieve and read a graph of passages.


Our approach first retrieves a passage graph by expanding a set of seed passages based on the graph structure of the knowledge base and the co-occurrence in text corpus (Figure~\ref{fig:intro}).
We then introduce a \reader\ model that extends \bert~\citep{bert} and propagates information from related passages and their relations, enabling knowledge-rich cross-passage representations.
Together, this approach allows for better coverage (e.g. the graph contains many passages that text-match retrieval would miss) and accuracy (e.g. by better combining information across related passages to find the best answer).

Experiments demonstrate significant improvements on three popular open-domain QA datasets: \webq~\citep{berant2013semantic}, \nq~\citep{kwiatkowski2019natural} and \triviaqa~\citep{joshi2017triviaqa}.  Our graph-based retrieval and reader models, together, improve accuracy consistently and significantly, outperforming the non-graph baselines by 2--11\% and matching or exceeding the \sota\ in every case without an expensive end-to-end training regime.
Through extensive ablations, we show that both graph-based retrieval and \reader\ models substantially contribute to the performance improvements, even when we fix the other component.

\section{Related Work}\label{sec:related}

\paragraph{Text-based Question Answering.}
Text-based open-domain QA is a long standing problem~\cite{voorhees1999trec,ferrucci2010building}. Recent work has focused on two-stage approaches that combine information retrieval with neural reading comprehension~\cite{chen2017reading,wang2018r,das2019multi,yang2019end}.
We follow this tradition but introduce a new framework which retrieves and reads a graph of passages.

Other graph retrieval methods have been developed, either using entity name matching~\cite{ding2019cognitive,xiong2019simple,godbole2019multi} or hyperlinks~\cite{asai2019learning}.
However, we are not aware of work integrating external knowledge bases or tightly coupling the approach with a graph reader, as we do in this paper.
Moreover, most previous graph-based approaches evaluate on questions that are explicitly written to encourage reasoning based on a chain of entities, such as \wikihop~\citep{wikihop} or \hotpot~\citep{hotpot}.
In this work, we instead focus on naturally gathered questions which require much more diverse types of cross paragraph reasoning.
\sewon{Updated.}

After retrieving evidence passages, most pipeline systems use a reading comprehension model to extract the answer.
Previous work either concatenates retrieved passages into a single sequence~\cite{neural-cascades,hotpot,song2018exploring} with no explicit model of how they are related, or reads each passage in parallel~\cite{clark2018multi,alberti2019bert,min2019compositional,wang2019multi} with no ability to fuse the information they contain.
To the best of our knowledge, reading passages by incorporating structural information across passages has not been studied previously.
The most related models are \citet{song2018exploring} and \citet{cao2019wikihop}, which fuse information through entities detected by entity linking and coreference resolution on \wikihop. In contrast, our model fuses information across passages, to better model the overall relationships between the different blocks of text.
\sewon{Last two sentence updated}

Other lines of research in open-domain QA include joint learning of retrieval and \reader\ components \citep{lee2019latent} or direct phrase retrieval in a large collection of documents~\cite{seo2019real}.
Although end-to-end training can further improve the performance of our approach, this paper only focuses on pipeline approaches since end-to-end training is computationally and memory expensive.


\paragraph{Knowledge Base Question Answering.}
Question answering over knowledge bases has also been well studied~\cite{berant2013semantic,kwiatkowski2013scaling,yih2015semantic}, typically without using any external text collections.
However, recent work has augmented knowledge bases with text from Wikipedia~\cite{das2017question,sun2018open,sun2019pullnet,xiong2019improving}, to increase factual coverage when a given knowledge base is incomplete.
In this paper, we study what can be loosely seen as an inverse problem. The model answers questions based on a large set of documents, and the knowledge base is used to better model relationships between different passages of text.


\section{Approach}\label{sec:model}

\begin{figure*}[tb]
\centering
\resizebox{1.85\columnwidth}{!}{\includegraphics[trim={0 25mm 24mm 0},clip]{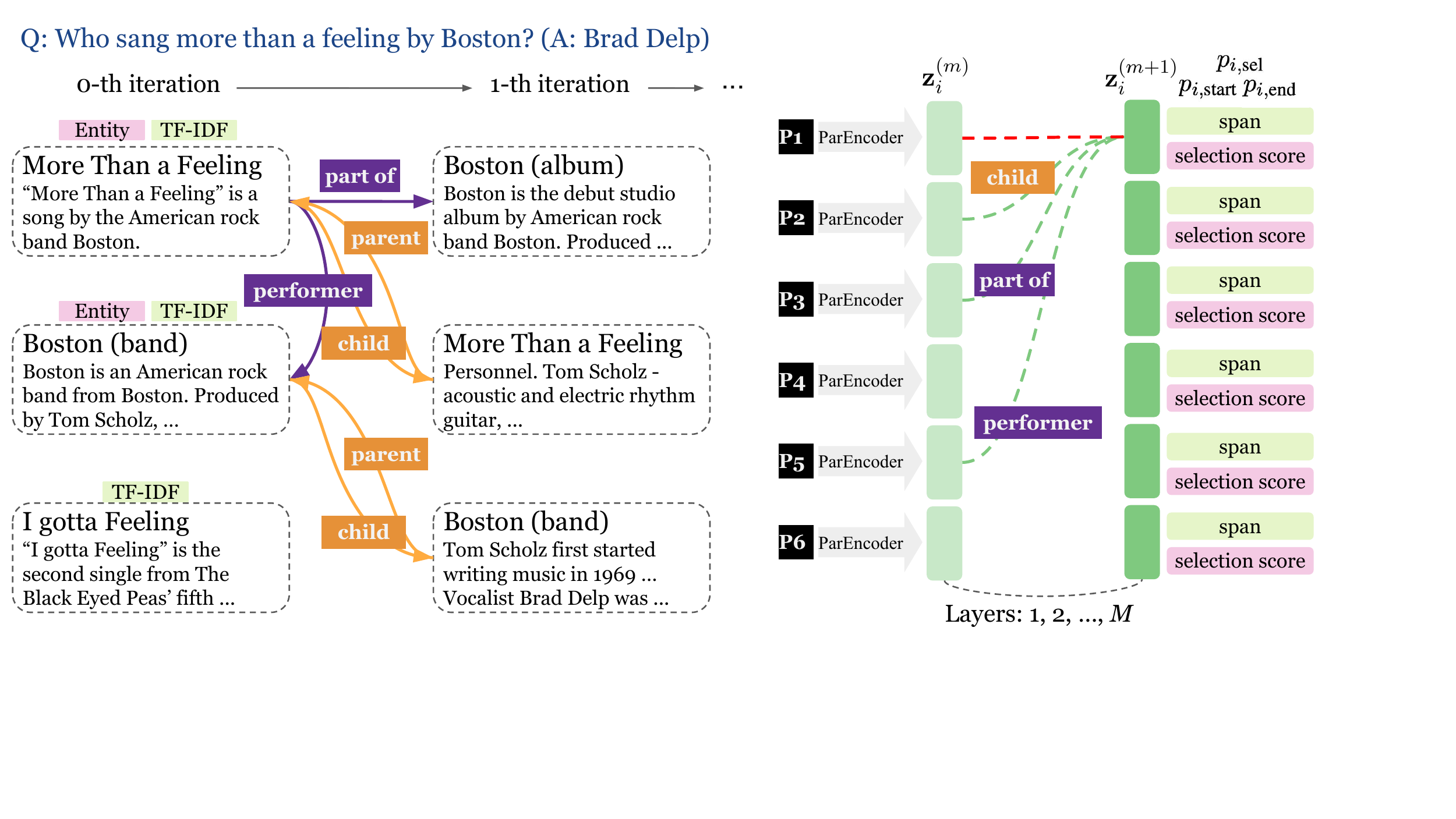}}
\vspace{-1em}
\caption{
    A diagram of our approach, consisting of \gretriever~(left) and \greader~(right).
    First, \gretriever\ constructs a graph of passages by obtaining seed passages through either entity linking or \tfidf, and expanding the graph based on Wikidata and Wikipedia (Section~\ref{subsec:model-retrieval}).
    \greader\ then takes this graph as an input, obtains initial passage representations, and updates them with respect to the graph, using $M$ fusion layers (Section~\ref{subsec:model-ours}).
}
\label{fig:model}
\end{figure*}

We present a new general approach for text-based open-domain question answering, which consists of a retrieval model \gretriever~and a neural \reader\ model \greader.
The overall approach is illustrated in Figure~\ref{fig:model}.
\gretriever\ retrieves a graph of passages in which vertices are passages and edges denote relationships between passages (Section~\ref{subsec:model-retrieval}).
\greader~ reads the input passage graph and returns the answer (Section~\ref{subsec:model-ours}).

\paragraph{Setup.}
The goal is to answer the question based on a text corpus $\mathcal{C}$, which consists of a large collection of articles and each of them can be divided into multiple passages.
We also assume an external knowledge base $\mathcal{K} = \{(e_1, r, e_2)\}$ exists where $e_1,e_2$ are entities and $r$ is a relation, and there is a 1-1 mapping between the KB entities and articles in the text corpus.
Specifically, we use Wikipedia as the text corpus $\mathcal{C}$ and Wikidata~\citep{vrandevcic2014wikidata} as the knowledge base $\mathcal{K}$, as there exists an alignment between the two resources and Wikipedia has been widely used before in open-domain question answering research~\cite{chen2017reading,seo2019real}.

\subsection{\gretriever}\label{subsec:model-retrieval}


Our retrieval approach takes a question as input and uses the knowledge base $\mathcal{K}$ to construct a graph of passages in $\mathcal{C}$.
It obtains seed passages and expands the passage graph through $M_\mathrm{ret}$ iterations, until it reaches the maximum number of passages $n$.
We denote $\mathcal{P}^{(m)}$ as the passages obtained in the $m$-th iteration, and describe how to obtain $\mathcal{P}^{(0)}$ (seed passages) and update the graph in the $m$-th iteration ($1 \leq m \leq M_\mathrm{ret}$). 




\vspace{-.2em}
\paragraph{Seed passages.} \gretriever\  starts with a set of Wikipedia articles by taking the union of (1) articles corresponding to the entities which are identified by an entity linking system~\citep{ferrucci2010building} on the input question; (2) the top $K_\mathrm{TFIDF}$ articles returned by a \tfidf\ based retrieval system. We choose the first passage of these articles as seed passages $\mathcal{P}^{(0)}$.


\paragraph{Graph expansion.}
Starting from seed passages $\mathcal{P}^{(0)}$, \gretriever\ expands the passage graph from $\mathcal{P}^{(m - 1)}$ to $\mathcal{P}^{(m)}$ by iterating over the following two methods, until it includes $n$ passages.

First, the passage graph is updated by adding passages that are related to $\mathcal{P}^{(m-1)}$ according to a relation present in Wikidata. 
Specifically, if $p_i \in \mathcal{P}^{(m-1)}$ and $p_j$ are the first passages of Wikipedia articles that correspond to KB entities $e_{p_i}$ and $e_{p_j}$ such that $(e_{p_i}, r_{i,j}, e_{p_j}) \in \mathcal{K}$, $p_j$ is added to the passage graph, being connected to $p_i$ through $r_{i,j}$.\footnote{
    Although this may include some entities that are not closely related to the question, it still increases the coverage of entities related to the answer, as shown in Section~\ref{subsec:exp-main-results}.
}
\sewon{This paragraph updated.}

Second, supporting passages for $\mathcal{P}^{(m-1)}$ are added to the passage graph.
Specifically, non-first passages of Wikipedia articles that are associated with $\mathcal{P}^{(m-1)}$ are ranked by BM25~\citep{robertson2009probabilistic} and the top $K_\mathrm{BM25}$ passages are chosen.
We construct relations between passages if they belong to the same Wikipedia article: $r_{i, j}$ is \child\ and $r_{j, i}$ is \parent\ if $p_i$ is the \textit{first passage} of the article and $p_j$ is another passage from the same article.

\paragraph{Final graph.} Finally, we retrieve a passage graph consisting of $n$ passages: $\{p_1, \dots, p_n\}$. The relations between the passages are denoted by $\{ r_{i,j} \mid 1 \leq i, j \leq n \}$, where $r_{i,j}$ is either a KB relation, \child, \parent\ or \none, indicating the relationship between a passage pair ($p_i$, $p_j$).

\subsection{\greader}\label{subsec:model-ours}

Our \greader~takes a question $q$ and $n$ retrieved passages $p_1, p_2, \ldots, p_n$ (and their relations $r_{i,j}$) and aims to output an answer to the question as a text span in one of retrieved passages. 
Instead of processing each passage independently, our approach obtains  knowledge-rich representations of passages by fusing information from linked passages across the graph structure. 


\subsubsection{Initial Passage Representation}
Formally, given the question $q$ and a passage $p_i$, \greader\ first obtains a question-aware passage representation:
\begin{equation*}
    \mathbf{P}_i = \mathrm{TextEncode}(q, p_i) \in \mathbb{R}^{L \times h},
\end{equation*}
where $L$ is the maximum length of each passage, and $h$ is the hidden dimension.
We use \bert~\citep{bert}, although the approach could be applied with many other encoders.

Additionally, \greader~encodes a relation $r_{i,j}$ through a relation encoder:
\begin{equation*}
    \mathbf{r}_{i,j} = \mathrm{RelationEncode}(r_{i,j}) \in \mathbb{R}^{h}.
\end{equation*}
We consider the most frequent 98 relations and group the other relations as \texttt{unk\_releation}, total to be 100 including \none.
We directly learn an an embedding matrix to get a vector representation for each relation, which works well in practice since we have relatively few relations and many examples of each.
\sewon{This paragraph updated.}

\subsubsection{Fusing Passage Representations}
\greader\ builds $M$ graph-aware fusion layers to update passage representations by propagating information through the edges of the graph, as depicted in the right side of Figure~\ref{fig:model}.
Specifically, \greader\ initializes passage representation with  $\mf{z}^{(0)}_{i} = \mathrm{MaxPool}(\mathbf{P}_i)$.
It then obtains new passage representation ${\mf{z}}^{(m)}_{i}$ for each fusion layer $1 \leq m \leq M$, based on its previous representation, all the adjacent passages and their relations.
Largely inspired by Graph Convolution Networks (GCN)~\citep{kipf2016semi,marcheggiani2017encoding}, we investigate two methods to obtain ${\mf{z}}^{(m)}_{i}$ from $\big\{ \mf{z}^{(m-1)}_j \big\}_{j=1}^n$ and $\big\{ \mf{r}_{i,j} \big\}_{j=1}^n$ as follows.


\paragraph{Binary.}
We first consider binary relations which encodes whether a passage pair is related or not, without incorporating relations. Specifically, \begin{eqnarray*}
    G_i &=& \{ j \mid  r_{i,j} \neq \none \}, \\
    {\mf{z}}^{(m+1)}_i &=& \frac{1}{|G_i|} \sum_{j \in G_i} \big( \mathbf{W}_f [{\mf{z}}^{(m)}_i \oplus \mf{z}^{(m)}_j] + \mf{b}_f \big),
\end{eqnarray*} where $\mathbf{W}_f$ and $\mf{b}_f$ are learnable parameters, and $\oplus$ is a concatenation.

\paragraph{Relation-aware.}
We then consider a relation-aware composition function: \begin{eqnarray*}
    {\mf{z}}^{(m+1)}_i &=& \frac{1}{n} \sum_{j=1}^n \big( \mathbf{W}_f [{\mf{z}}^{(m)}_i \oplus f(\mf{r}_{i,j}, \mf{z}^{(m)}_j)] + \mf{b}_f \big),
\end{eqnarray*} where $f$ is a composition function, $\oplus$ is a concatenation, and $\mathbf{W}_f$ and $\mf{b}_f$ are learnable parameters.
We use concatenation for the composition function,
$f(\mf{r}_{i,j}, \mf{z}^{(m)}_j) = \textsc{Concat}(\mf{r}_{i,j}, \mf{z}^{(m)}_j)$, for simplicity because it worked as well as more complex functions such as element-wise multiplication and bilinear mappings in our early experiments.
\sewon{Last sentence updated, and cut the ablation in Section 5.1.} 





\subsubsection{Answering Questions}
\greader\ uses the updated passage representations $\mf{z}^{(M)}_1, \ldots, \mf{z}^{(M)}_n$ to compute the probability of $p_i$ being an evidence passage.
Denote $\mathbf{Z} = [\mf{z}^{(M)}_1, \ldots, \mf{z}^{(M)}_n] \in \mathbb{R}^{h \times n}$, we define \begin{eqnarray*}
    P_\textrm{sel}(i)  =  \mathrm{softmax} \big( \mathbf{Z}^{\intercal} \mf{w}_\mathrm{sel}  \big)_i,
\end{eqnarray*}
where $\mf{w}_\mathrm{sel} \in \mathbb{R}^{h}$ is a learnable vector.
Once the evidence passage is chosen by $i^* = \argmax_{1 \leq i \leq n} P_\textrm{sel}(i)$, the probability of a span $1 \leq j \leq k \leq L$ in the passage $p_{i^*}$ being an answer is computed as $P_{\textrm{start},i^*}(j) \times P_{\textrm{end},i^*}(k)$, where \begin{eqnarray*}
    P_{\textrm{start},i^*}(j)  &=& \mathrm{softmax} \big( \mathbf{P}_{i^*} \mf{w}_\textrm{start}\big)_j, \\
    P_{\textrm{end},i^*}(k) &=& \mathrm{softmax} \big( \mathbf{P}_{i^*} \mf{w}_\textrm{end}\big)_k,
\end{eqnarray*}
where $\mf{w}_\textrm{start}, \mf{w}_\textrm{end} \in \mathbb{R}^{h}$ are learnable vectors.


For training, we use the maximum marginal likelihood objective by maximizing: \begin{eqnarray*} &&\sum_{i=1}^n \mathbb{I}[|\mathcal{S}_i|>0] \mathrm{log} P_\textrm{sel}(i) \\
    &+& \sum_{i=1}^n \mathrm{log} \left(\sum\limits_{(s_i,e_i) \in \mathcal{S}_i}  \left( P_{\textrm{start},i}(s_i) \times  P_{\textrm{end},i}(e_i) \right) \right)
\end{eqnarray*} where $\mathcal{S}_i$ is a set of spans which correspond to the answer text in $p_i$. 
We tried using $\mathbf{Z}$ for span predictions, but did not see meaningful improvements.
We hypothesize it is because span prediction given the correct passage is an easier task compared to choosing the right evidence passage.\footnote{We observed that over 80\% of the error cases the baseline model made are due to the incorrect passage selection on all datasets.}

\section{Experiments}\label{sec:exp}

\begin{table}
    \centering
    \footnotesize
    \setlength\tabcolsep{3.8pt}
    \begin{tabular}{lrrrrrr}
    \toprule
         \tf{Dataset} &  \multicolumn{3}{c}{\tf{Statistics}} &  \multicolumn{3}{c}{\tf{Graph density}} \\
         & Train & Dev & Test & Cross & Inner & Total \\
    \midrule
        \textsc{WebQ} &  3417 & 361 & 2032 & 1.53 & 0.88 & 2.41 \\
        \textsc{NaturalQ} & 79168 & 8757 & 3610 & 1.26 & 0.90 & 2.16 \\
        \triviaqa & 78785 & 8837 & 11313 & 1.28 & 0.88 & 2.16 \\
    \bottomrule
    \end{tabular}

    \caption{Dataset statistics and density of the graph (\# of relations per a passage). Cross, Inner and Total denote cross-document relations (KB relations), inner-document relations (\child, \parent) and their sum.
    }
    \label{tab:dataset}
\end{table}

\begin{table*}
\centering
\small
\begin{tabular}{l l  P{0.8cm}P{0.8cm} P{0.8cm}P{0.8cm} P{0.8cm}P{0.8cm}}
\toprule
    \tf{Retriever} & \tf{\readercap} &  \multicolumn{2}{c}{\tf{\webq}} & \multicolumn{2}{c}{\tf{\nq}} & \multicolumn{2}{c}{\tf{\triviaqa}} \\
    & & Dev & Test & Dev & Test & Dev & Test \\
\midrule
        Text-match & \parallelqa            & 23.6 & 25.2 & 26.1 & 25.8 & 52.1 & 52.1 \\
        Text-match & \parallelqabatch       & 19.9 & 20.8 & 28.9 & 28.7 & 54.5 & 54.0 \\
\midrule
        \gretriever & \parallelqa           & \webqpardev & \webqpartest & \nqpardev & \nqpartest & \triviaqapardev & \triviaqapartest \\
        \gretriever & \parallelqabatch      & \webqbatchdev & \webqbatchtest & \nqbatchdev & \nqbatchtest & \triviaqabatchdev & \triviaqabatchtest \\
        \gretriever & \greader\ (binary)    & \webqbinarydev & \textbf{\webqbinarytest} & \nqbinarydev & \nqbinarytest & \triviaqabinarydev & \triviaqabinarytest \\
        \gretriever & \greader\ (relation)  & \webqreldev & \webqreltest & \textbf{\nqreldev} & \textbf{\nqreltest} & \textbf{\triviaqareldev} & \textbf{\triviaqareltest} \\
\midrule
        \multicolumn{2}{l}{Previous best (pipeline)}    & - & 18.5$^a$ & 31.7$^b$ & 32.6$^b$ & {50.7}$^c$ & {50.9}$^c$ \\
        \multicolumn{2}{l}{Previous best (end-to-end)} & \textbf{38.5}$^d$ & \textbf{36.4}$^d$ & {31.3}$^d$ & {33.3}$^d$ & 45.1$^d$ & 45.0$^d$ \\
\bottomrule
\end{tabular}
\caption{
    Overall results on the \dev\ and the test set of three datasets.
    We also report the previous best results, both with pipeline and end-to-end: $^a$\citet{lin2018denoising}, $^b$\citet{asai2019learning}, $^c$\citet{min2019discrete}, $^d$\citet{lee2019latent}.
    Note that the development sets used in \citet{lin2018denoising} and \citet{lee2019latent} are slightly different but the test sets are the same; \citet{asai2019learning} uses a better pretrained model (a whole word masking \bert$_\text{LARGE}$).
} \label{tab:results}
\end{table*}

\subsection{Datasets}\label{subsec:exp-dataset}

We evaluate our model on three open-domain question answering datasets, where the evaluation metric is Exact Match.
(1) \textbf{\webq}~\citep{berant2013semantic} is originally a QA dataset designed to answer questions based on Freebase; the questions were collected through Google Suggest API.
We follow~\citet{chen2017reading} and frame the problem as a span selection task over Wikipedia.
(2) \textbf{\nq}~\citep{kwiatkowski2019natural} consists of questions collected using the Google search engine; questions with short answers up to 5 tokens are taken following~\citet{lee2019latent}.
(3) \textbf{\triviaqa}~\citep{joshi2017triviaqa} consists of questions from trivia and quiz-league websites.
For all datasets, we only use question and answer pairs for training and testing, and discard the provided evidence documents which are part of reading comprehension tasks. We follow the data splits from \citet{chen2017reading} for \webq\ and \citet{min2019discrete} on \nq\ and \triviaqa.\footnote{\url{https://bit.ly/2q8mshc} and \url{https://bit.ly/2HK1Fqn}.} Table~\ref{tab:dataset} shows the statistics of the datasets and the density of the graph (number of relations per passage) retrieved by \gretriever.


\subsection{Baselines}\label{subsec:exp-baseline}
For retrieval, we compare our \gretriever~to a pure text-match based retrieval method which retrieves Wikipedia articles based on \tfidf\ scores~\citep{chen2017reading} and ranks their passages through BM25~\citep{robertson2009probabilistic}.
This is to investigate if leveraging the knowledge base actually improves the retrieval component.

For reader, we compare our \greader\ with two competitive baselines which read each passage in parallel, \textbf{\parallelqa} and \textbf{\parallelqabatch}. Both baselines obtain question-aware passage representations $\mathbf{P}_1, \dots, \mathbf{P}_n$ as described in Section~\ref{subsec:model-ours} with a different way of calculating $P_\textrm{sel}(i)$.
\textbf{\parallelqa} computes $P_\textrm{sel}(i)$ using a binary classifier: 
\begin{equation*}
    P_\textrm{sel}(i) = \mathrm{softmax} \big( \mathbf{W}_\mathrm{sel} \mathrm{maxpool}(\mathbf{P}_i) \big)_1,
\end{equation*} where $\mathbf{W}_\mathrm{sel} \in \mathbb{R}^{2 \times h}$ is a learnable matrix~\citep{alberti2019bert,min2019compositional}.
\textbf{\parallelqabatch} is similar to \parallelqa\ but takes a softmax across passages, inspired by \citet{clark2018multi}\footnote{Our preliminary result indicates this variant slightly outperforms the original model, S-Norm.}: 
\begin{equation*}
    P_\textrm{sel}(i) = \mathrm{softmax} \big( \mathbf{\hat{P}}^{\intercal} \mf{w}_\mathrm{sel}  \big)_i,
\end{equation*} where $\mathbf{\hat{P}} = [\mathrm{MaxPool}(\mathbf{P}_1), \ldots, \mathrm{MaxPool}(\mathbf{P}_n)] \in \mathbb{R}^{h \times n}$ and $\mf{w}_\mathrm{sel} \in \mathbb{R}^{h}$ is learned.
Note that \parallelqabatch\ is exactly the same as \greader\ with no fusion layer.
Finally, for both baselines, the probability of the span is computed in the same way as described in Section~\ref{subsec:model-ours}.

\subsection{Implementation Details}\label{subsec:exp-details}

We use the Wikipedia dump from 2018-12-20 and the Wikidata dump from 2019-06-01\footnote{\url{archive.org/download/enwiki-20181220} and \url{dumps.wikimedia.org/Wikidatawiki/20190601}}.
We use TAGME~\cite{tagme}\footnote{\url{github.com/gammaliu/tagme}} as an entity linking system.
We split the article into passages with natural breaks and merge consecutive ones with up to a maximum length of 300 tokens.
We use $K_\mathrm{TFIDF}=5, K_\mathrm{BM25}=40$ for \webq\ and $K_\mathrm{TFIDF}=10, K_\mathrm{BM25}=80$ for the rest two, which empirically set $M_\mathrm{ret}$ as $2$ and $1$, respectively.

We use the uncased version of \bert$_\text{BASE}$~\citep{bert} for question-aware passage representations $\mathrm{TextEncode}$.
When training \greader, we cannot feed the full passage graph in the same batch due to the memory constraints.
Therefore, for every parameter update, we sample at most $20$ passages where one of them contains the answer text, either uniformly at random or by taking a subgraph.
For each model, we experiment with fusion layers of $M=\{1, 2, 3\}$ and two sampling methods, and choose the number that gives the best result on the \dev\ set. 
More details can be found in Appendix~\ref{app:details}.

\subsection{Main Results}\label{subsec:exp-main-results}

\paragraph{Model Comparisions}
The main results are given in Table~\ref{tab:results}. We observed three overall trends:
\begin{inparaenum}[(1)]
    \item \gretriever\ offers significant performance gains over text-match retrieval when we compare within the same \reader\ across all datasets, e.g., 1--11\% absolute gains with \parallelqabatch. This indicates that graph-based retrieval provides passages with significantly better evidence to answer the question.
    \item \greader\ outperforms two \parallelqa\ baselines consistently across all datasets, achieving 1--5\% absolute gains. This result demonstrates that fusing information across passages is more effective than reading each passage in isolation.
    \item \greader\ using relations offers some improvement over \greader\ with binary relations. The gains are smaller than expected, likely because the relations are inferred based from the text.\footnote{
        In order to verify this hypothesis, we modify our reader to have an output layer for relation classification, and observe that the accuracy is over 80\% for all datasets. 
    }
\end{inparaenum}

\paragraph{State-of-the-art results.} We also compare our results to the previous best models, both pipeline and end-to-end approaches for open-doman QA, in Table~\ref{tab:results}.
Our best-performing model outperforms previously published pipeline models by 6--18\%, showcasing the benefit of our graph retrieval and reader models.
In particular, our models with \gretriever\ (both baseline and \greader) outperform the previous best graph-based retrieval model~\citep{asai2019learning}\footnote{
    \sewon{Updated.}
    To the best of our knowledge, \citet{asai2019learning} (1) is the only graph-based approach evaluated on naturally found questions and (2) also outperforms other graph-based approaches on \hotpot.
} by a large margin, despite the fact that they used a stronger BERT model than ours as the base model.
Our model also outperforms or matches the end-to-end model~\citep{lee2019latent} which is expensive to train as it uses an extra pretraining strategy.
Although not explored in this paper, our framework can be trained end-to-end as well, which has a great potential to further advance the \sota.

\section{Analyses}\label{sec:analysis}To better understand model performance, we report a number of ablation studies (Section~\ref{subsec:exp-ablations}) and a qualitative analysis (Section~\ref{subsec:qualitative}).


\subsection{Ablation Studies}\label{subsec:exp-ablations}


\begin{table}[!t]
    \centering
    \small
    \setlength\tabcolsep{4pt}
    \begin{tabular}{l c c}
    \toprule
        Retriever & \tf{WebQ} & \tf{Natural Q} \\
    \midrule
    \textbf{\em Reader: \parallelqabatch} \\
    Text-match          & 19.9 & 28.9 \\
    \oldretrieval       & \webqscbatch & \nqscbatch \\
    \gretriever         & \textbf{\webqbatchdev} & \textbf{\nqbatchdev} \\
    \midrule
    \textbf{\em Reader: \greader~(binary)} \\
    \oldretrieval       & \webqscbinary & \nqscbinary \\
    \gretriever         & \textbf{\webqbinarydev} & \textbf{\nqbinarydev} \\
    \bottomrule
    \end{tabular}
    \caption{
    {\bf Effect of different retrieval methods}. We compare text-match retrieval and \greader\ with `\oldretrieval' (described in Section~\ref{subsec:exp-ablations}).
}
\label{tab:retrieval-ablations}
\end{table}

\paragraph{Effect of different retrieval methods.}
Table~\ref{tab:retrieval-ablations} compares text-match retrieval and \gretriever\ with `\oldretrieval', a variant of \gretriever\ where we take the union of text-match retrieval and Wikidata-based retrieval, each computed in isolation.
Specifically, the text-match retrieval is the baseline described in Section~\ref{subsec:exp-baseline}, and Wikidata-based retrieval is done by obtaining seed passages through entity linking and updating the passage graph only through Wikidata.
This variant can be seen as {\it late} combination between the text-match and Wikidata-based retrieval, whereas \gretriever\ provides {\it early} combination.
Although `\oldretrieval' outperforms text-match retrieval by a large margin, our \gretriever\ significantly outperforms this method, showing the importance of jointly leveraging text-match and Wikidata for graph construction.

\begin{table}[tb]
    \centering
    \small
    \begin{tabular}{l c c}
    \toprule
         &  \tf{\webq} & \tf{Natural Q} \\
    \midrule
        Fully connected     & \webqfc  & \nqfc \\
        Empty               & \webqempty  & \nqempty \\
    \midrule
        Cross-doc           & \textbf{\webqcross}  & \nqcross \\
        Inner-doc           & \webqinner  & \nqinner \\
        Cross+Inner         & \webqbinarydev  & \textbf{\nqbinarydev} \\
    \bottomrule
    \end{tabular}
    \caption{
        \textbf{Effects of different relation types in \greader}.
        We compare input graphs containing different sets of edges, where we use \gretriever\ and \greader~(binary).
    }\label{tab:relation-types}
\end{table}

\paragraph{Effect of different relation types in \greader.}
Table~\ref{tab:relation-types} compares the effect of using different relation types in the constructed graph of passages for \greader, showing results for the following settings: (a) {\it fully connected,} which connects all pairs of passages, (b) {\it empty}, which does not include any edges between passages, (c) {\it cross-doc}, which only includes edges between passages according to the Wikidata relations, (d)  {\it inner-doc}, which only includes \child\ and \parent, and (e) {\it cross+inner}, which includes both cross-doc and inner-doc, corresponding to the graph constructed by our approach. 
Results indicate that cross-doc and inner-doc relations achieve good performance across two datasets.
In particular, using relation information is better than ignoring relation information (fully connected, empty), demonstrating the importance of selecting a good set of graph edges.
\sewon{This paragraph briefly updated to emphasize that it is ablation on the reader.}

\begin{table}[!t]
    \centering
    \small
    \setlength\tabcolsep{4pt}
    \begin{tabular}{l c c}
    \toprule
         & \tf{WebQ} & \tf{Natural Q} \\
    \midrule
    \parallelqabatch        & \webqbatchdev & \nqbatchdev \\
    \parallelqabatch~(pairs from graph) & \webqpair & \nqpair \\
    \parallelqabatch~(all pairs) & \webqpairall & \nqpairall \\
    \greader~(binary)       & \textbf{\webqbinarydev} & \nqbinarydev \\
    \greader~(relation)     & \textbf{\webqreldev} & \textbf{\nqreldev} \\
    \bottomrule
    \end{tabular}
    \caption{
    {\bf Comparison to passage concatenation}. For all rows, \gretriever\ is used.}
\label{tab:concatenation}
\end{table}

\paragraph{Comparison to passage concatenation.}\label{subsec:ablations-pair-baselines}
Table~\ref{tab:concatenation} compares the performance of our graph-based method with two baseline readers where a concatenation of passage pairs is included as input, and \parallelqabatch\ reads each of them in isolation.
First, \parallelqabatch~(pairs from graph) concatenates passage pairs that are related in the input graph, along with the relation text.
Second, \parallelqabatch~(all pairs) concatenates all passage pairs. 
For these baselines, the concatenated passages are up to 300 tokens.\footnote{
    We split each passage up to 145 tokens and set the relation text to be up to 10 tokens.
}
For \parallelqabatch~(pairs from graph), we use $n=20$ instead of $n=\{40,80\}$.\footnote{This restrition is needed because there are too many passage pairs: even $n=20$ gives $20+190=210$ passages.} It is worth noting that concatenating more passages into a single input is non-trivial due to the fixed input length of \bert. Details are provided in Appendix~\ref{app:details}.
Results show that concatenating passages is not competitive, potentially because truncating each passage causes significant information loss.


\begin{figure*}[tb]
\centering
\resizebox{2\columnwidth}{!}{\includegraphics[width=\columnwidth]{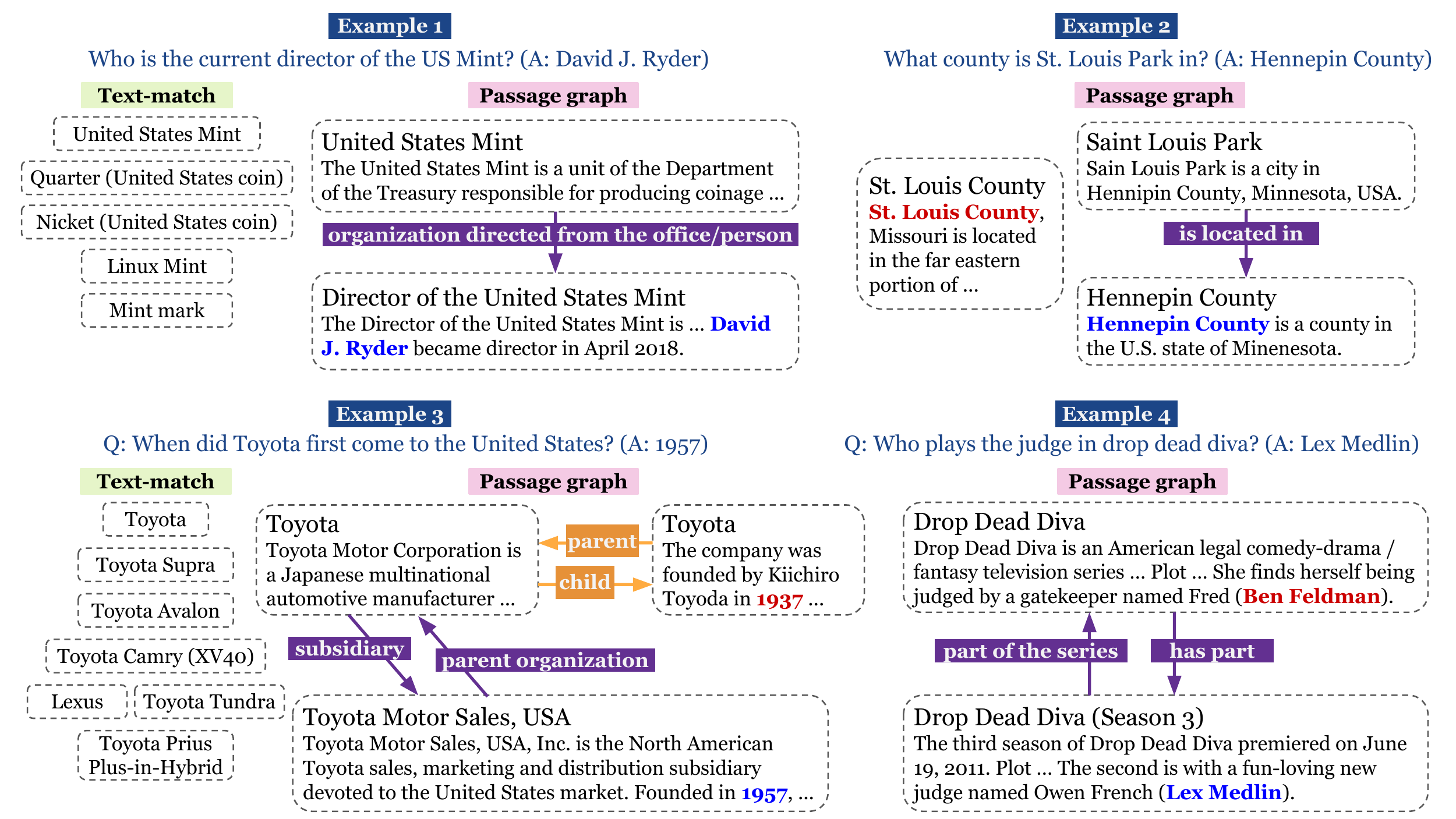}}
\caption{
    Examples from \nq\ and \webq\ where predictions from \parallelqabatch\ and \greader\ (both with \gretriever) are denoted by \redtext{red} and \bluetext{blue} text, respectively.
    Subsets of the retrieved graphs are reported. 
    Detailed analyses in Section~\ref{subsec:qualitative}. 
}
\label{fig:analyses}
\end{figure*}

\subsection{Qualitative Results}
\label{subsec:qualitative}

Figure~\ref{fig:analyses} shows a few examples from \nq\ and \webq. Appendix~\ref{app:analyses} lists additional examples.
They include cases where our method incorporates knowledge-rich relationships between passages to find the correct evidence and answer the question. 

\paragraph{Knowledge helps retrieving evidence passages.}
In Example 1, text-match retrieval does not retrieve the article `Director of the United States Mint' and fails to retrieve any passage about the director of the US Mint. However, \gretriever\ retrieves the correct evidence passage by using the relationship between `United States Mint' and `Director of the United States Mint', enabling \greader\ to successfully predict the answer.
Similarly, in Example 3, Wikidata enables \gretriever\ to retrieve `Toyota Motor Sales, USA' which contains the evidence to the question, whereas text-match retrieval fails to do so.\footnote{For both Example 1 and 3, initial retrieved articles include some passages containing the evidence, but BM25 passage ranking misses them.}

\paragraph{Relation information explicitly supports the answer.}
Although Example 2 appears to be easy to humans since the passage from `Saint Louis Park' alone provides the enough evidence, \parallelqabatch\ with no relation information makes a wrong prediction, `St. Louis County', potentially because of the similarity in names.
However, Wikidata relation in Example 2, {\em is located in}, explicitly supports the evidence to answer the question, therefore, \greader\ which leverages graph information easily predicts the right answer.

\paragraph{Relation information enables the model to synthesize across related passages.}

In Example 3, \parallelqabatch\ makes a wrong prediction from the passage `Toyota', potentially because this passage seems more related to the company. \greader, however, leverages the relationship between `Toyota' and `Toyota Motor Sales, USA', and predicts the correct answer. Similarly in Example 4, \parallelqabatch\ predicts ``Ben Feldman" as an answer potentially due to the word ``judged by". However, leveraging relations in the graph {\em has part} and {\em part of the series}, \greader\ infers that two passages belong to the same series and `Drop Dead Diva (Season 3)' mentions the judge more explicitly.


\section{Conclusion}\label{sec:concl}
We proposed a general approach for text-based open-domain question answering that integrates graph structure at every stage to construct, retrieve and read a graph of passages.
Our retrieval method leverages both text corpus and a knowledge base to find a relevant set of passages and their relations. Our reader then propagates information according to the input graph, enabling knowledge-rich cross-passage representations.
Our approach consistently outperforms competitive baselines on three open-domain QA datasets, \webq, \nq\ and \triviaqa. We also included a detailed qualitative analysis to illustrate which components contribute the most to the overall system performance.



\bibliography{journal-abbr,bib}
\bibliographystyle{acl_natbib}
\clearpage
\appendix

\section{Training details}\label{app:details}

\subsection{Details \& Hyperparameters}
All experiments are done in Python 3.5 and PyTorch 1.1.0~\citep{pytorch}.
For \bert, we use the uncased version of \bert$_\text{BASE}$ and \textsc{pytorch-transformers}~\citep{Wolf2019HuggingFacesTS}\footnote{\url{github.com/huggingface/transformers}}.
Specifically, given a question $Q$ and a passage $P_i$ where the title of the originated article is $T_i$, we form a sequence $S_i = Q:{\tt [SEP]}:{\tt <t>}:T_i:{\tt </t>}:P_i$, where : indicates a concatenation and ${\tt [SEP]}$ is a special token. This sequence is then fed into \bert\ and the hidden representation of the sequence from the last layer is chosen as a question-aware passage representation.
For the embedding matrix for the relation encoder, we keep 100 relations (\none, \texttt{UNK} and top 98 relations), which cover over 95\% of all relations on all datasets.

For \parallelqa, we use a batch size of $10$ on \webq\ and $60$ on \nq\ and \triviaqa. For \parallelqabatch\ and \greader, we use a batch size of $8$ on \webq\ and $16$ on the rest two.
For each fusion layer, we apply dropout~\citep{srivastava2014dropout} with a probability of $0.3$.
For training, we evaluate the model on the development set periodically, and stop training when Exact Match score does not improve $10$ times.
For all other hyperparameters not mentioned, we follow the default settings from \textsc{pytorch-transformers}.

As mentioned in Section~\ref{subsec:exp-details}, for each model, we experiment with $M=\{1, 2, 3\}$ and two sampling methods, and choose the number that gives the best result on the \dev\ set. The chosen hyperparameters for each model is reported in Table~\ref{tab:hyperparameters}.

For inference, we experiment with the number of input passages $n=\{40,80\}$ and choose the best one on the development set for testing.
We restrict the predicted span to be a Freebase entity string on \webq, following \citet{chen2017reading}.

\begin{table}[th!]
    \centering \small
    \begin{tabular}{l l c c}
    \toprule
        Dataset & \greader & $M$ & sample \\
    \midrule
        & binary & 1 & U \\
        & relation, concat & 1 & U \\
        \webq & relation, elm-wise & 1 & G\\
        & relation, bilinear & 1 & U \\
    \midrule
        & binary & 1 & G \\
        & relation, concat & 3 & G \\
        \textsc{Natural Q.} & relation, elm-wise & 1 & U \\
        & relation, bilinear & 2 & G \\
    \midrule
        & binary & 1 & G \\
        & relation, concat & 1 & G \\
        \triviaqa & relation, elm-wise & 1 & U \\
        & relation, bilinear & 2 & U \\
    \bottomrule
    \end{tabular}
\caption{
    Hyperparameters used for experiments.
    $M$ denotes the number of fusion layers, and `sample' denotes a sampling method for training, where `U' and `G' indicate an uniform sampling and a subgraph sampling, respectively.
}
\label{tab:hyperparameters}
\vspace{-8pt}
\end{table}



\subsection{Details for baselines with passage concatenation.}
We design two baselines which concatenate a passage pair.

First, \parallelqabatch~(pairs from graph) concatenates passage pairs that are related in the input graph, along with the relation text. Specifically, if $p_i$ and $p_j$ are connected through $r_{i,j}$, all of $p_i$, $p_j$, $p_i:{\tt{[SEP]}}:r_{i,j}:{\tt{[SEP]}}:p_j$ are included as input passages of \parallelqabatch. We limit the length of the passages and the relation text to be 145 and 10, respectively, so that the total length to be up to $300$. The number of the final input passages to the model will be $n + |\{r_{i,j} | r_{i,j} \neq \text{none}\}|$, where $0 \leq |\{r_{i,j} | r_{i,j} \neq \text{none}\}| \leq \frac{n(n-1)}{2}$ (but typically much smaller than $\frac{n(n-1)}{2}$ as the input graph is very sparse).

Similarly, \parallelqabatch~(all pairs) concatenate all passage pairs. For all $1 \leq i < j \leq n$, $p_i$, $p_j$, $p_i:{\tt{[SEP]}}:p_j$ are included as input passages of \parallelqabatch. As $p_i$ and $p_j$ may not have a relation, the relation text $r_{i,j}$ is omitted as an input. Again, the length limit for $p_i$ and $p_j$ is 145. The number of the final input passages to the model will be $n + \frac{n(n-1)}{2}$. As there are too many input passages for this baseline, we use $n=20$ instead of $n=\{40,80\}$.

\begin{figure*}[tb]
\centering
\resizebox{2\columnwidth}{!}{\includegraphics[width=\columnwidth]{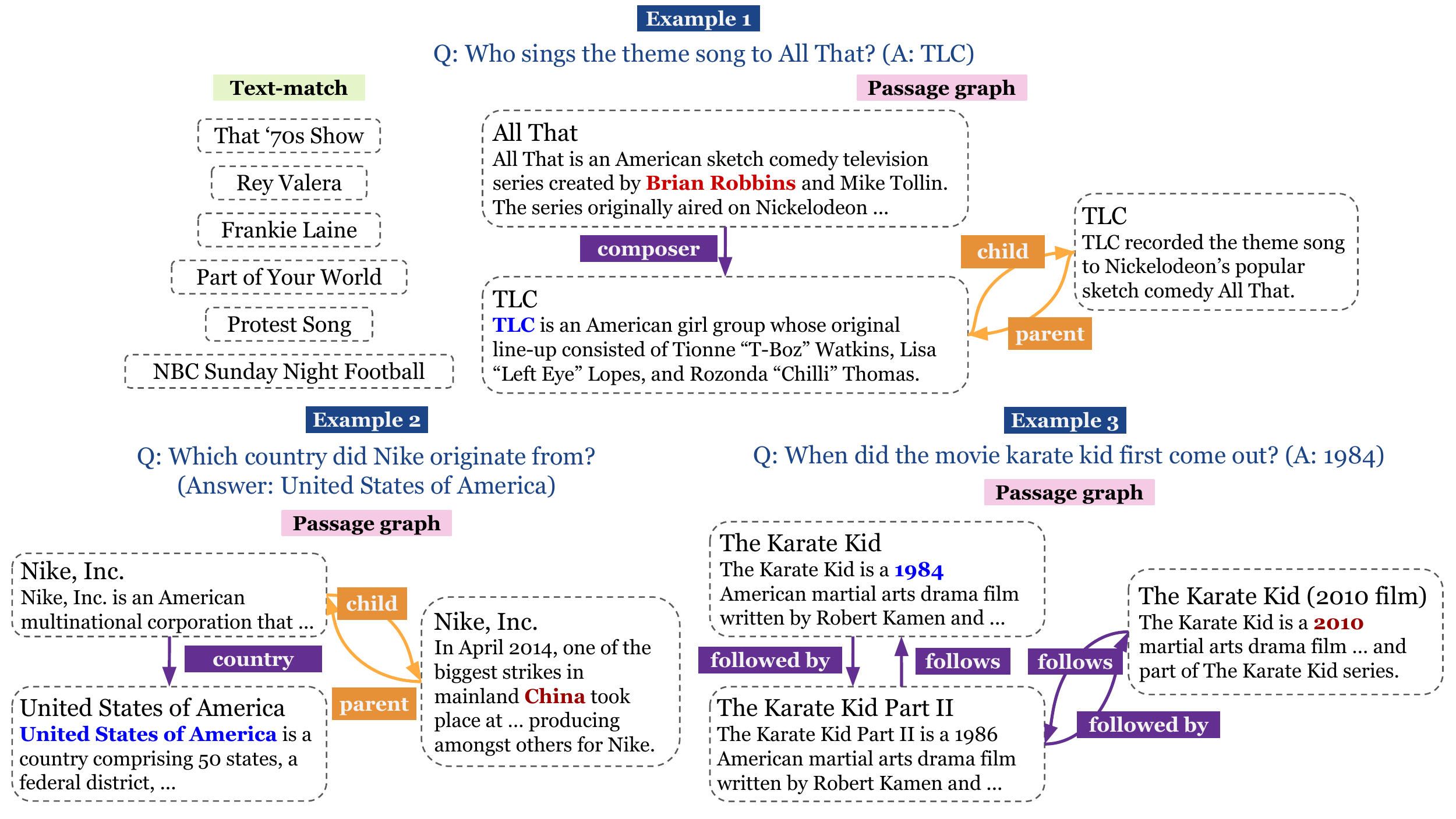}}
\caption{
    Examples from \nq\ and \webq\ where predictions from \parallelqabatch\ and \greader\ (both with \gretriever) are denoted by \redtext{red} and \bluetext{blue} text, respectively.
    Subsets of graph information are reported. 
    Detailed analyses in Section~\ref{app:analyses}.
}
\label{fig:analyses-more}
\end{figure*}

For both baselines, as the limit for a single passage is different from other baselines (145 vs. 300), we run the retrieval again by following the same method but just split the article into passages with a different length limit.

\section{More qualitative analyses}\label{app:analyses}

Figure~\ref{fig:analyses-more} depict more examples where our model predicts the correct answer. We describe how our model outperforms baselines for each of retrieval and reading component.

\paragraph{Retrieval through entity linking.} In Example 1, text-match retrieval fails to retrieve the evidence passages because it fails to capture `All That' as a key entity.
\gretriever, on the other hand, retrieves the article ``All That" by entity linking. 
It is worth to note that this was particularly common, when the key entities are composed of common words that \tfidf\ does not capture its importance, e.g., ``Who plays letty in \underline{bring it on all or nothing}?" or ``Who sings \underline{does he love you} with Reba?".

\paragraph{Retrieval through knowledge.}
In Example 1, entity linking was not enough for evidence to answer the question, because the article ``All That" does not contain the singer of the theme song. Meanwhile, WikiData contains a triple $<$``All That", {\em composer}, ``TLC"$>$, allowing the retrieval of the passage ``TLC".

\paragraph{Reading by synthesizing across passages.}
In Example 1, although \gretriever\ retrieves the evidence passage, \parallelqabatch\ which does not leverage the graph information predicts the wrong span by choosing the first person name in the passage from ``All That". \greader, however, leverages the relation {\em composer} and predicts the correct answer.

In Example 2, although the question appears to by easy for humans, \parallelqabatch\ retrieves ``China" as an answer, potentially because it is the only country name mentioned in retrieve passages from the article ``Nike, Inc." However, \greader\ leverages the relation {\em country} and predicts the correct answer.

Example 3 requires to reason across multiple passages, as it asks about the first advent of the movie series that have similar titles. \parallelqabatch, which reads each passage in isolation, predicts ``2010" from the wrong passage. \greader, however, incorporates relations {\em followed by} and {\em follows} and successfully distinguishes the first movie.

\end{document}